\documentclass{article} 
\usepackage[T1]{fontenc}
\usepackage[utf8]{inputenc}
\usepackage{iclr2026_conference,times}

\usepackage{amsmath,amsfonts,bm}

\def\eqref#1{equation~\ref{#1}}

\def\1{\bm{1}}

\DeclareMathAlphabet{\mathsfit}{\encodingdefault}{\sfdefault}{m}{sl}
\SetMathAlphabet{\mathsfit}{bold}{\encodingdefault}{\sfdefault}{bx}{n}

\usepackage[hidelinks]{hyperref}
\hypersetup{pdftitle={DWM: Separating World Effects from Actions in Latent World Models},pdfauthor={Yi-Ge Zhang, Tianqi Du, Qi Zhang, Yisen Wang}}
\usepackage{url}
\usepackage{booktabs}
\usepackage{amsmath}
\usepackage{amssymb}
\usepackage{pifont}
\usepackage{xcolor}
\definecolor{revisionpurple}{HTML}{7030A0}
\definecolor{revisionorange}{HTML}{D97706}
\usepackage{graphicx}
\usepackage{wrapfig}
\usepackage{microtype}
\usepackage{ragged2e}

\title{DWM: Separating World Effects from Actions in Latent World Models}

\author{Yi-Ge Zhang \quad Tianqi Du \quad Qi Zhang \quad Yisen Wang\thanks{Corresponding author: Yisen Wang (yisen.wang@pku.edu.cn)} \\
Peking University
}

\iclrfinalcopy 
\begin{document}

\maketitle
 \begin{abstract}                                              
Latent world models provide an effective way to model environment dynamics for control, but existing action-conditioned formulations typically supervise the next-latent transition using a single prediction target, requiring the model to explain all sources of state change through one unified learning signal. In real-world environments, however, state transitions often arise from two distinct sources: changes caused by the agent's actions and changes induced by the environment's intrinsic dynamics. In this paper, we introduce DWM (Decomposed World Model), a supervision-level framework that explicitly separates these two sources of transition. The action-invariant source can occur even under a null action and includes effects such as gravity-driven sliding, inertia, contact rebound, and persistent drift. Modeling these two sources jointly can entangle their effects in the learned latent transition, making it difficult to attribute observed changes to their underlying causes and limiting the transferability of the learned dynamics. DWM augments a latent world model with an auxiliary world head that captures action-invariant changes and is regularized by a normalized world-contrastive objective. The original prediction head is coupled with the world head through an orthogonality constraint, encouraging the predicted transition to decompose into an action-invariant component and a complementary action-driven component. This formulation requires no modification to the underlying world-model architecture or inference pipeline. To evaluate DWM in environments with persistent world effects, we construct W-variants of three standard control benchmarks, PushT-W, Reacher-W, and TwoRoom-W, each introducing a distinct form of action-invariant dynamics. Across multiple representative latent world models, DWM consistently improves both prediction quality and downstream control performance on these benchmarks.

    \end{abstract}
                     
  \section{Introduction}                                                                                              
    \label{sec:intro}                                                                                                   
                                                                                                                        
Latent world models have gained increasing attention for model-based planning in learned latent spaces~\citep{hafner2019planet,hafner2020dreamer,hafner2021dreamerv2,hafner2023dreamerv3,hansen2022tdmpc,hansen2024tdmpc2}. Within this family, joint-embedding predictive architectures~\citep{assran2023ijepa,bardes2024vjepa} predict future content directly in a learned latent space rather than reconstructing raw signals, offering compact and semantically meaningful representations for downstream control. Recent works such as LeWM~\citep{maes2026lewm} and DINO-WM~\citep{zhou2024dino} learn action-conditioned dynamics directly between latent representations: an encoder maps observations to a latent representation, and an action-conditioned predictor is trained to regress the next-step latent, so that the entire latent transition is supervised by a single next-latent target. These predicted latents are then used for goal-conditioned planning across a range of control tasks, where both methods achieve strong planning performance on standard manipulation and navigation benchmarks.
  
A physical transition, however, is rarely caused by the agent's action alone. Consider a block sliding on a tilted surface. Part of its motion is produced by the agent's push, while another part would occur even if the agent did nothing, because gravity continues to move the block. We refer to these two sources of change as the \textbf{action effect} and the \textbf{world effect}, respectively. More precisely, the world effect is the part of the transition that would persist under the same state and history if the current action were replaced by a null action, and the action effect is the remaining change attributable to the chosen action. Similar decompositions arise when objects move because of inertia, articulated bodies swing under gravity, or agents drift under persistent external forces.

In current representative methods such as LeWM and DINO-WM, the world effect and the action effect are conflated into a single next-latent target, without any explicit signal to separate them. As a result, the model struggles to isolate what the action actually does, ultimately degrading downstream planning performance. Figure~\ref{fig:teaser} illustrates this issue on PushT-W, a gravity-perturbed variant of the standard PushT task. Under a zero action, the block continues to slide along the direction of gravity. A conventional single-head latent world model fails to reproduce this passive motion accurately, and its predicted trajectory consequently deviates from the goal. These observations motivate the central question of this work:


\begin{figure}[!t]
\centering
\includegraphics[width=0.85\linewidth]{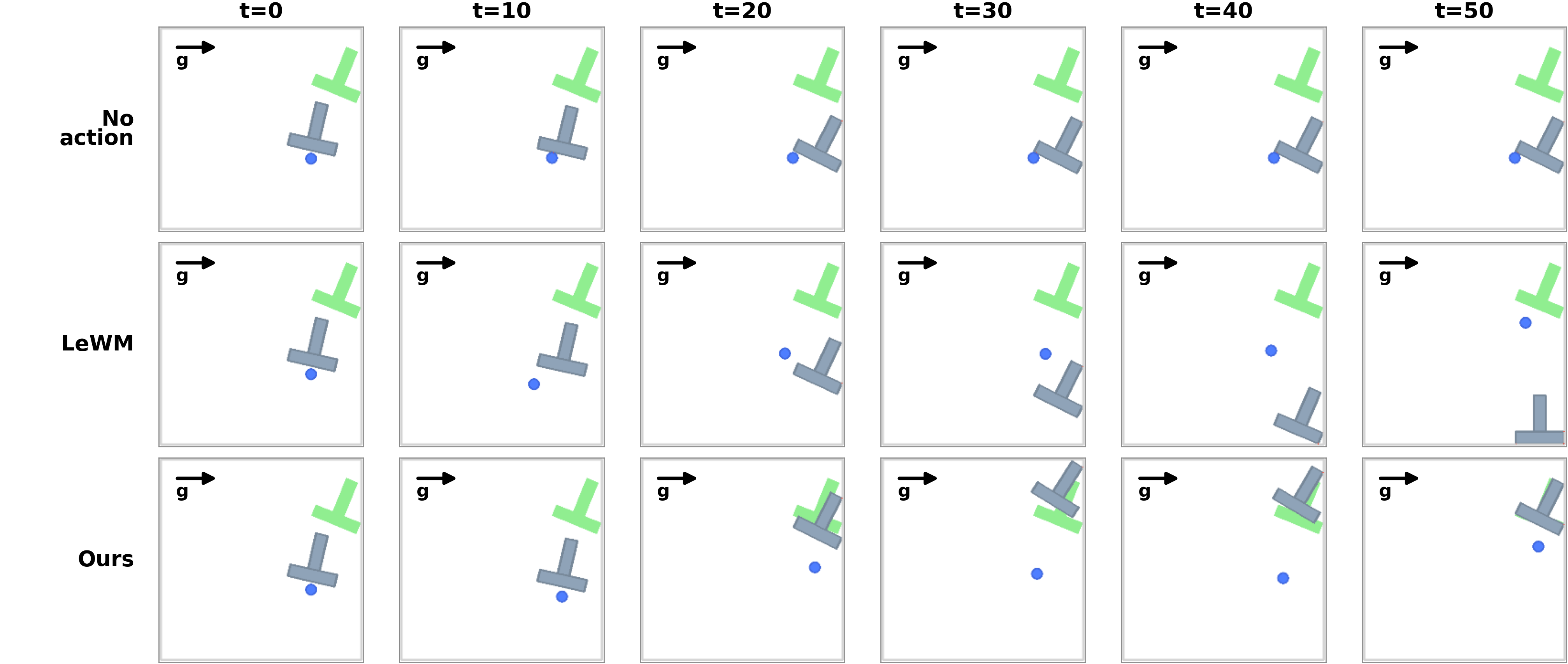}
\caption{World/action entanglement on PushT-W. The
\textcolor{gray}{gray T} is the current block, the
\textcolor{green!60!black}{green T} is the goal, and the
\textcolor{blue}{blue circle} is the pusher; the arrow $\mathbf{g}$
marks the persistent gravity that defines the world effect.
\textbf{Top:} under a zero action, the block slides along $\mathbf{g}$,
isolating the world effect.
\textbf{Middle:} the single-head LeWM baseline misses this passive
motion and drifts away from the goal.
\textbf{Bottom:} DWM disentangles the two components and brings the
block onto the goal.}
\label{fig:teaser}
\end{figure}

     \begin{quote}                                                                                                       
    \emph{Can latent world models be improved by explicitly separating what the world does on its own from what the agent's action changes?}                                                                           
    \end{quote}                                                           
      

We introduce Decomposed World Model (DWM), a world/action-decomposed training framework for action-conditioned latent world models. During training, DWM adds an auxiliary world head alongside the original prediction head; by perturbing the current action while keeping the history fixed, we require the world-head prediction to remain stable across actions yet discriminative across states, so it captures the action-invariant component of the transition. We then decompose the complete prediction into the world-head prediction and its complementary action-driven component, and add an orthogonality regularizer to encourage the two components to encode complementary transition information. Because the auxiliary branch is used only during training, DWM leaves the inference-time architecture and planning pipeline unchanged, and can be integrated with existing latent world models.

To evaluate DWM, we construct three controlled benchmarks, PushT-W, Reacher-W, and TwoRoom-W, by augmenting the original PushT, Reacher, and TwoRoom tasks with gravity-driven sliding, vertical-plane gravity, and constant environmental drift, respectively. We further evaluate DWM on Ball-in-Cup, a more challenging environment in which gravity, string tension, and ball-cup contact jointly give rise to a compound world effect. Across the three W-variants, integrating DWM with LeWM and DINO-WM yields average absolute improvements of $13.1\%$ and $10.2\%$ in CEM planning success, respectively. DWM also retains its advantage under OOD gravity ($+14\%$ on PushT-W), improves planning on Ball-in-Cup, and consistently reduces the autoregressive rollout error across all tasks. Our main contributions are as follows:

    
    \begin{itemize}
                                                     
        \item We identify a supervision-level limitation of joint-embedding latent world models trained against a single next-latent target: this target merges changes caused by the agent with changes that would occur without the current action, providing no explicit signal for separating the two sources of transition.
        \item We introduce a controlled benchmark suite consisting of PushT-W, Reacher-W, and TwoRoom-W, which adds persistent action-invariant dynamics to standard control tasks while retaining their original objectives and flat counterparts as controls.
        \item We propose DWM, a unified training framework whose auxiliary world head and world/action separation objectives can be integrated with representative latent world models while preserving the original inference pipeline.
        \item We demonstrate that DWM substantially outperforms the corresponding baselines when integrated with different representative latent world models, and across multiple realistic settings involving diverse world effects.                                                                      \end{itemize}

\section{World and Action Effects In Latent Dynamics}

\subsection{Preliminaries}
\label{sec:prelim}

\paragraph{Action-conditioned latent world model.}
\label{sec:prelim:lewm}

We consider action-conditioned latent world models such as LeWM~\citep{maes2026lewm} and DINO-WM~\citep{zhou2024dino}, together with their recent variants~\citep{gao2026fastlewm,huang2026trajlewm,zhao2026subjepa,thil2026sdjepa,wang2026adajepa,du2026vlwm}. These models operate on trajectories of raw observations $o_t$ (e.g., pixels) and executed actions $a_t$, and share three components:
\begin{itemize}
    \item an {encoder} $\phi$ that maps each raw observation
    $o_t$ to a compact latent state $z_t=\phi(o_t)$, i.e., the
    representation on which all downstream prediction and planning
    operates;
    \item an {action-conditioned predictor} $g$ that summarizes a
    short context window of past latents and actions
    $(z_{t-h+1:t},\,a_{t-h+1:t})$ into an intermediate rollout state
    $r_t = g(z_{t-h+1:t},\,a_{t-h+1:t})$, where $h$ is the context
    length; $r_t$ can be read as ``what the model believes about the
    world after observing history up to $t$ and taking $a_t$'';
    \item a {projection head} that reads out $r_t$ and produces
    the predicted next latent $\hat z_{t+1}$\footnote{We denote the generic one-step prediction $\hat z_{t+1}$ simply by
$\hat z$ when no ambiguity arises, while retaining full time indices
for multi-step rollouts.}.
\end{itemize}
The predictor is trained by regressing $\hat z_{t+1}$ towards the
true next latent $z_{t+1}$ with a mean-squared loss
$\mathcal{L}_{\mathrm{pred}}$.
To prevent the learned representation from collapsing, some methods such as LeWM additionally introduce
a latent-space regularizer $\mathcal{L}_{\mathrm{reg}}$ (instantiated
as SIGReg in LeWM). We write the base training objective as
$\mathcal{L}_{\mathrm{WM}}=\mathcal{L}_{\mathrm{pred}}+\lambda_{\mathrm{reg}}\mathcal{L}_{\mathrm{reg}}$,
where the regularizer term is optional.


\paragraph{CEM planning in latent space.}
\label{sec:prelim:cem}

For planning, we use model-predictive planning with the Cross-Entropy
Method (CEM) directly in the latent space induced by the encoder
$\phi$ and the predictor $g$~\citep{rubinstein1997cem,chua2018pets}.
Given a raw image goal $o_g$, the goal is embedded once into a
latent target $z_g=\phi(o_g)$, and planning is carried out entirely
by rolling the world model forward from the current context. Concretely,
at each planning step the current history
$(z_{t-h+1:t},\,a_{t-h+1:t-1})$ is fixed, and CEM optimizes an
action sequence $a_{t:t+H-1}$ of horizon $H$ under a Gaussian
proposal $\mathcal{N}(\mu_i,\Sigma_i)$ that is refined over
iterations $i=1,\dots,I$:
\begin{equation}
\{a^{(k)}_{t:t+H-1}\}_{k=1}^{N} \sim \mathcal{N}(\mu_i,\Sigma_i),
\qquad
\hat z^{(k)}_{t+j+1}
=
\pi\!\left(
g\!\left(\hat z^{(k)}_{t+j-h+1:t+j},\,
      a^{(k)}_{t+j-h+1:t+j}\right)
\right),
\label{eq:cem-rollout}
\end{equation}
where $\pi$ denotes the projection head reading out $r_t$ (in
Section~\ref{sec:method}, $\pi=h_p$), and $\hat z^{(k)}_{t}=z_t$
initializes the rollout with the encoded context. Each candidate
sequence is scored by a cost function
$c(\hat z^{(k)}_{t+1:t+H};z_g)$ that we instantiate either as a
terminal latent distance
\begin{equation}
c\!\left(\hat z^{(k)}_{t+1:t+H};z_g\right)
\;=\;
\lVert \hat z^{(k)}_{t+H}-z_g\rVert_2^2
\label{eq:cem-cost-latent}
\end{equation}
for tasks where the raw latent distance is informative, or as
$c(\hat z^{(k)}_{t+1:t+H};z_g)=-s(\hat z^{(k)}_{t+1:t+H};z_g)$ for
Ball-in-Cup (Section~\ref{sec:results:hard}), with $s$ a learned
success head that scores the entire predicted rollout. The elite
set $\mathcal{E}_i$ of the top-$K$ lowest-cost samples is then used
to update the proposal,
\begin{equation}
\mu_{i+1}
=
\frac{1}{K}\sum_{k\in\mathcal{E}_i} a^{(k)}_{t:t+H-1},
\qquad
\Sigma_{i+1}
=
\frac{1}{K}\sum_{k\in\mathcal{E}_i}
\bigl(a^{(k)}_{t:t+H-1}-\mu_{i+1}\bigr)^{\!2},
\label{eq:cem-update}
\end{equation}
and the first action of the resulting mean, $\mu_I[0]$, is executed
in the environment; the context is then advanced and planning
repeats in receding-horizon fashion.

\subsection{Defining World and Action Effects}
\label{sec:effects}

The introduction described the world effect as the part of a transition
that would persist if the current action were replaced by a null action.
We now formalize this counterfactual notion in latent space. Fixing the
same history $z_{t-h+1:t}$ and past actions $a_{t-h+1:t-1}$, but
replacing the current action $a_t$ with the zero action $\mathbf{0}$,
we define the action-invariant component of the latent transition as
\begin{equation}
\Delta z^{\mathrm{world}}_t
\;:=\;
\mathbb{E}\!\left[\,z_{t+1}\;\bigl|\;
z_{t-h+1:t},\,a_{t-h+1:t-1},\,a_t{=}\mathbf{0}\,\right]
\;-\; z_t.
\label{eq:def-world}
\end{equation}
Intuitively, $\Delta z^{\mathrm{world}}_t$ is what the world would do
on its own if the agent chose not to intervene at step $t$. It captures
the persistent phenomena that our benchmarks are designed to expose,
such as gravity-driven sliding, inertia, and drift. Crucially,
Eq.~\ref{eq:def-world} depends only on the environment dynamics and
the current context, not on the actual $a_t$, so
$\Delta z^{\mathrm{world}}_t$ is well-defined for every transition
regardless of whether the two effects combine additively in the true
dynamics.

Given $\Delta z^{\mathrm{world}}_t$, we define the action-driven
component as the residual,
$\Delta z^{\mathrm{action}}_t
\;:=\; \Delta z_t \, -\, \Delta z^{\mathrm{world}}_t$, so that
\begin{equation}
\Delta z_t \;=\; \Delta z^{\mathrm{world}}_t \;+\; \Delta z^{\mathrm{action}}_t
\label{eq:decomp}
\end{equation}
holds by construction. This is a modelling identity rather than a claim
that the two effects combine additively at the physical level. In real
environments they may couple nonlinearly, for instance when the same
push produces different net motion depending on the current
gravity-induced velocity. Our formulation does not assume such coupling
away; instead, the residual absorbs whatever interaction the two effects
have, and the interaction itself is later carried by the shared latent
representation that both prediction paths read from
(Section~\ref{sec:method:arch}).

\subsection{Diagnosing Monolithic Transition Supervision}
\label{sec:motivation}

LeWM~\citep{maes2026lewm} and DINO-WM~\citep{zhou2024dino} report strong visual planning performance, such as on standard PushT, but these standard tasks contain little to no world effect, whereas world effects are pervasive in the real world. To probe prediction and planning under this more realistic setting, we introduce {PushT-W}, a gravity-perturbed variant of the standard 2D pushing task in which the block keeps sliding under gravity irrespective of the action (its full construction is deferred to Section~\ref{sec:tasks}), and use LeWM, a recent representative method, for our analysis.

\paragraph{Monolithic supervision degrades under a world effect.}
\label{sec:mot:blind}

On standard PushT, where the block barely moves without the agent's
input and the transition is essentially action-driven, LeWM under the
standard single-target supervision reaches a CEM planning success
rate of $94.0\%$ (Appendix~\ref{app:flat-suite}). On PushT-W, gravity
continually drags the block regardless of the action, so a substantial
share of each transition is action-invariant; the same model, trained
to convergence, achieves only $32\%$. This sharp drop indicates that
motion produced by gravity is folded into the same next-latent target
as action-driven motion, and the resulting predictor is not accurate
enough to support CEM planning. We next examine whether this
shortfall can be attributed to the training data.

\paragraph{Injecting zero-action data does not separate world from action.}
\label{sec:mot:data}

A natural hypothesis is that the failure above stems from insufficient
exposure to purely action-invariant transitions: if the training set
rarely contains frames in which the block moves under gravity alone,
the model may never see the world effect in isolation. To test this,
we augment the PushT-W training set with a $30\%$ mixture of
zero-action frames and retrain the same single-target model. Despite
explicit exposure to pure gravity-driven transitions, the retrained
model attains a CEM planning success rate of $30\%$, statistically
indistinguishable from the $32\%$ baseline. The bottleneck therefore
lies not in data coverage but in the training objective itself.

\section{DWM: World/Action Disentanglement for Latent World Models}
\label{sec:method}

We build on the additive identity
$\Delta z_t=\Delta z^{\mathrm{world}}_t+\Delta z^{\mathrm{action}}_t$
(Section~\ref{sec:effects}) as an inductive bias, and reallocate the
transition supervision of a latent world model so that its two sources
are learned along separate prediction paths
(Section~\ref{sec:method:arch}), each shaped by its own auxiliary
objective (Section~\ref{sec:method:loss}). This construction
uses the components that action-conditioned latent world models
typically share---an encoder $\varphi$, an action-conditioned predictor
$g$ that produces a rollout state $r_t$, and a readout of the predicted
next latent---so it can be integrated with such models without altering
their architecture or inference.

\subsection{Decomposing the Predicted Transition}
\label{sec:method:arch}

\begin{figure}[t]
    \centering
    \includegraphics[width=0.82\linewidth]{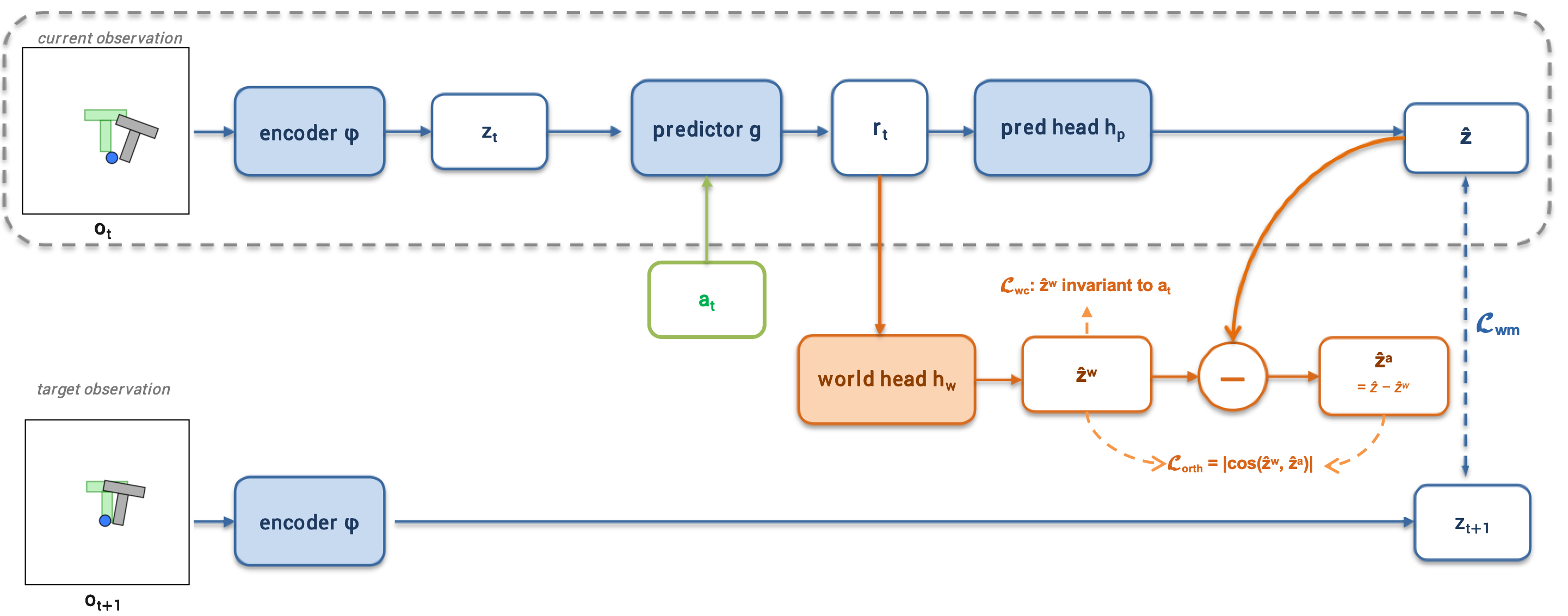}
    \caption{\textbf{DWM implementation.}
    DWM preserves the original prediction pathway of an action-conditioned latent world model and adds an auxiliary world head $h_w$ during training. The world head produces $\hat z^{\,w}$, while the complementary action-driven component is defined as $\hat z^{\,a}=\hat z-\hat z^{\,w}$. The auxiliary objectives in Section~\ref{sec:method:loss} shape the semantics of the two components. The world head is removed at inference, leaving the original prediction pipeline unchanged.}
    \label{fig:architecture}
\end{figure}

To instantiate the decomposition in Eq.~\ref{eq:decomp} without changing the original world-model prediction pathway, DWM preserves the prediction of the full next latent and introduces an auxiliary branch to isolate the world-driven component. As illustrated in Figure~\ref{fig:architecture}, given the action-conditioned predictor state $r_t$, the original prediction head $h_p$ continues to produce the full prediction $\hat z=h_p(r_t)$ and is supervised by the standard world-model objective $\mathcal{L}_{\mathrm{WM}}$. In parallel, we attach a lightweight world head $h_w$ to the same predictor state and obtain $\hat z^{\,w}=h_w(r_t)$. We then define the remaining action-driven component as the residual between the full prediction and the world component:
\begin{equation}
    \hat z^{\,w}=h_w(r_t), \qquad
    \hat z^{\,a}=\hat z-\hat z^{\,w}, \qquad
    \hat z=\hat z^{\,w}+\hat z^{\,a}.
\end{equation}
This construction converts the additive transition decomposition in Eq.~\ref{eq:decomp} into an explicit decomposition of the model prediction. Rather than learning two independent predictors, DWM retains the original prediction as the complete next-state estimate and uses the auxiliary branch to determine how this prediction is partitioned into world-driven and action-driven components. This residual formulation also guarantees that the two components always reconstruct the original prediction exactly.

Importantly, the decomposition is used only to shape representation learning during training. At inference, the auxiliary world head and the decomposition are discarded, and the original prediction head $h_p$ remains the sole output of the world model. DWM therefore introduces no additional inference-time computation and can be applied without modifying the deployment pipeline of the underlying latent world model.
\subsection{Learning Complementary World and Action Components}
\label{sec:method:loss}

The parameterization above alone does not guarantee that the two
components correspond to world-driven and action-driven changes. We
therefore introduce two complementary objectives to identify the
desired decomposition. The original world-model loss
$\mathcal{L}_{\mathrm{WM}}$ is retained to supervise the full
prediction $\hat z$, while the additional losses constrain
$\hat z^{\,w}$ to be action-invariant and separate it from the
action-driven residual:
\begin{equation}
\mathcal{L}
=
\mathcal{L}_{\mathrm{WM}}
+
\lambda_{\mathrm{wc}}\mathcal{L}_{\mathrm{wc}}
+
\lambda_{\mathrm{orth}}\mathcal{L}_{\mathrm{orth}}.
\label{eq:method_loss}
\end{equation}

\paragraph{World-contrastive objective.}
To make $\hat z^{\,w}$ invariant to the current action while retaining
state-specific information, we construct two views of the same context
using different actions. For each context $i$ in a mini-batch of size
$B$, we keep the observation history fixed and replace the current
action $a_t^{(i)}$ with a perturbed action $\tilde a_t^{(i)}$, obtained
in practice by randomly permuting actions within the batch. Passing
the two action-conditioned predictor states through the shared world
head produces $\hat z^{\,w,(i)}$ and
$\tilde{\hat z}^{\,w,(i)}$. After $\ell_2$ normalization, denoted by
$u^{(i)}$ and $\tilde u^{(i)}$, we treat the two representations of
the same context as a positive pair and representations from different
contexts as negatives:
\begin{equation}
\mathcal{L}_{\mathrm{wc}}
= -\frac{1}{2B}\sum_{i=1}^{B}\!\left[
    \log \frac{e^{u^{(i)\top}\tilde u^{(i)}/\tau}}
               {\sum_{j} e^{u^{(i)\top}\tilde u^{(j)}/\tau}}
  + \log \frac{e^{\tilde u^{(i)\top}u^{(i)}/\tau}}
               {\sum_{j} e^{\tilde u^{(i)\top}u^{(j)}/\tau}}
\right].
\label{eq:wc}
\end{equation}
This objective encourages the world component to remain unchanged
under different actions, while the contrastive negatives prevent the
representation from collapsing across states.

\paragraph{Orthogonality objective.}
The world-contrastive objective encourages the world component to be invariant to the current action, but action invariance alone does not guarantee a clean decomposition. Since the full prediction is represented as the sum of the world component and the residual, the two components may still encode overlapping information about the same state transition. In particular, the residual may redundantly retain directions already captured by the world head, making the separation ambiguous. We therefore introduce an orthogonality constraint, $L_{orth} = |cos(\hat{z}^w, \hat{z}^a)|$, which discourages shared directions between the two components. Combined with the world-contrastive objective, this encourages $\hat{z}^w$ to capture the action-invariant part of the transition, while pushing the residual $\hat{z}^a$ to explain complementary variation needed for the action-conditioned prediction. The original world-model loss simultaneously ensures that their sum remains predictive of the full next latent.
Together, $\mathcal{L}_{\mathrm{wc}}$ identifies the action-invariant
part of the transition, while $\mathcal{L}_{\mathrm{orth}}$ encourages
the residual to capture complementary action-dependent variation.
The original $\mathcal{L}_{\mathrm{WM}}$ simultaneously preserves the
predictive quality of their sum.

\section{Experiments}
\label{sec:experiments}

We begin by describing the benchmarks and experimental
setup (Section~\ref{sec:tasks}), and then evaluate DWM on the three
W-variants, comparing it against representative latent world models and
disentanglement methods (Section~\ref{sec:results:main}). To understand
where these gains come from, we analyze the learned world/action
decomposition (Section~\ref{sec:results:evidence}) and evaluate
multi-step prediction and generalization to unseen gravity directions
(Section~\ref{sec:results:multistep-ood}). We further test DWM under multiple
interacting world effects on Ball-in-Cup (Section~\ref{sec:results:hard}),
and finally show that DWM generalizes across distinct latent world models
(Section~\ref{sec:results:dinowm}).

\subsection{Benchmarks and Experimental Setup}
\label{sec:tasks}

To evaluate world/action disentanglement under
controlled changes in persistent autonomous motion, we pair
each of three standard tasks---PushT, Reacher, and
TwoRoom---with a \emph{W-variant} that introduces a persistent
action-invariant source of motion (gravity-driven sliding,
vertical-plane gravity, or constant environmental drift,
respectively). Each W-variant keeps the same goal as its flat
counterpart but adds a source of motion that persists even when the
agent is idle; the flat counterpart serves as a comparability
control. We further evaluate on Ball-in-Cup as a more
demanding simulated task with multiple interacting world effects.
Tossing and catching the ball require predicting motion shaped by
gravity, string tension, and ball--cup contact, whose combined
influence changes throughout the trajectory.
Full physical parameterizations, dataset sizes, and OOD splits are
deferred to Appendix~\ref{app:tasks} (Table~\ref{tab:tasks}).


\paragraph{Model and optimization.}
\label{sec:setup:model}

We evaluate DWM on action-conditioned latent world models built from an
encoder $\varphi$ and a Transformer predictor $g$ (Section~\ref{sec:prelim:lewm}),
using a context length $h\!=\!3$ and observations subsampled at a stride
of $5$ frames. DWM attaches the world head $h_w$
(Section~\ref{sec:method:arch}) in parallel to the pred head $h_p$ as a
two-layer MLP with BatchNorm, leaving the encoder and predictor
otherwise unchanged. All trainable modules are optimized jointly with
AdamW, and the number of training epochs is fixed per task family before
evaluation, so that a base model and its DWM variant are compared under
an identical training and planning budget. We instantiate this recipe on
two representative models---LeWM and DINO-WM---whose full architecture,
optimization, and hardware settings are given in
Appendix~\ref{app:impl}.
\par

\paragraph{Evaluation metrics.}
\label{sec:setup:metrics}

\begin{itemize}
    \item \textbf{Prediction quality}: one-step \texttt{pred\_loss}
    between $\hat z_{t+1}$ and $z_{t+1}$, reported both offline and
    under autoregressive rollout. Latent-space MSE for PushT-W and
    TwoRoom-W; qpos error (rad) for Reacher-W.
    \item \textbf{Disentanglement diagnostics}: response of the world
    and pred heads under controlled action perturbations, quantifying
    whether the world head is action-invariant while the pred head
    remains action-sensitive.
    \item \textbf{OOD gravity}: on PushT-W, cosine similarity of
    $\hat z^{\,a}$ across held-out gravities, and CEM success at
    gravities distinct from training.
    \item \textbf{Planning}: CEM success rate on a fixed set of
    start--goal pairs, averaged over $3$ planner seeds; per-task
    protocols (goal offset, evaluation budget, CEM sample and
    refinement counts, success criterion) are in
    Appendix~\ref{app:tasks}.
\end{itemize}

\subsection{Main Results: Planning Under World Effects}
\label{sec:results:main}


\begin{figure}[h]
\centering
\includegraphics[width=0.72\linewidth]{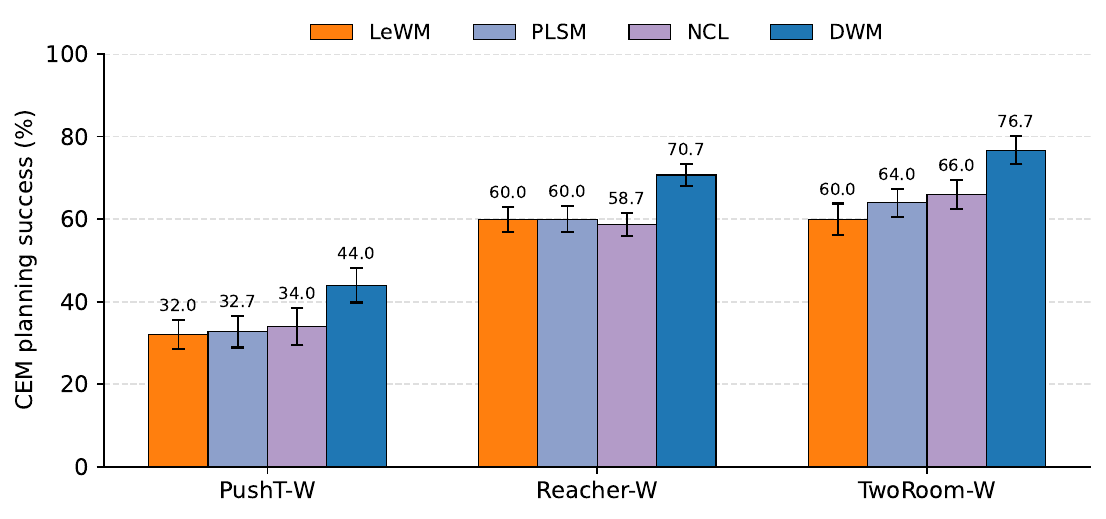}
\caption{CEM planning success of LeWM, PLSM, NCL, and DWM on the
three W-variants. Bars report the mean over $3$ CEM planner seeds
with $n{=}50$ start--goal pairs per seed; error bars denote the
across-seed standard deviation.}
\label{fig:main-flat}
\end{figure}

%
%
%
%

We compare DWM against LeWM and two
representative disentanglement objectives adapted on top of it:
PLSM~\citep{saanum2024softly}, which minimizes the mutual information
between the current latent state and the action-induced latent change,
and NCL~\citep{wang2024ncl}, which imposes non-negativity constraints
on contrastive representations. Both baselines share LeWM's single
next-latent target and neither reallocates supervision between world-
and action-driven components. We evaluate all methods
under one training recipe (Section~\ref{sec:setup:model}) and one CEM
protocol ($3$ planner seeds, $n\!=\!50$ start--goal pairs per seed;
per-task configurations in Appendix~\ref{app:tasks}).
Figure~\ref{fig:main-flat} reports CEM planning
success on the three W-variant tasks. LeWM drops from $94.0\%$,
$82.0\%$, and $94.0\%$ on the flat counterparts
(Appendix~\ref{app:flat-suite}) to $32.0\%$, $60.0\%$, and $60.0\%$
once a persistent world effect is present, indicating that
single-target supervision struggles in this regime; PLSM and NCL
remain close to it at $32.7\%$, $60.0\%$, $64.0\%$ and
$34.0\%$, $58.7\%$, $66.0\%$ on PushT-W, Reacher-W, and TwoRoom-W.

DWM supervises an action-invariant world
component and its complementary action-driven component, raising CEM
planning success on the three tasks from $32.0\%$, $60.0\%$, $60.0\%$
to $44.0\%$, $70.7\%$, and $76.7\%$, an average absolute improvement of
$13.1$ percentage points over the baseline;
Appendix~\ref{app:ablation} shows that both $\mathcal{L}_{\mathrm{wc}}$
and $\mathcal{L}_{\mathrm{orth}}$ are needed for this gain, and
Appendix~\ref{app:flat-suite} confirms that the separation does not
degrade performance on the flat PushT, Reacher, and TwoRoom tasks.

\subsection{Does DWM Learn the Intended Decomposition?}
\label{sec:results:evidence}

Building on the planning gains of Section~\ref{sec:results:main}, we
next examine whether the intended world/action decomposition is
realized in the learned representation. Specifically, for each
validation window we fix the observation history $z_{t-h+1:t}$ and
sample $K$ alternative actions $\{\tilde a_t^{(k)}\}_{k=1}^{K}$; we
then measure the response of each head by the variance and
peak-to-peak range of its outputs across these perturbations, which
capture the average and worst-case sensitivity, respectively. A
successful decomposition should preserve the pred head's sensitivity
to the action while rendering the world head invariant to it.

\begin{table}[h]
\caption{Disentanglement diagnostics on the three W-variant tasks.
For each validation window the current action $a_t$ is perturbed with
$K$ random alternatives while the history is held fixed; we report the
per-sample variance and peak-to-peak range of each head across these
perturbations, averaged over windows. Since LeWM has no world head,
only the pred-head variance column is populated for it.}
\label{tab:disentangle}
\begin{center}
\small
\setlength{\tabcolsep}{4pt}
\begin{tabular}{lccccc}
\toprule
\textbf{Task}
 & \textbf{Pred-head var}
 & \textbf{Pred-head var}
 & \textbf{World-head var}
 & \textbf{World/pred}
 & \textbf{World/pred} \\
 & (LeWM)
 & (DWM)
 & (DWM)
 & \textbf{var ratio $\downarrow$}
 & \textbf{range ratio $\downarrow$} \\
\midrule
PushT-W    & $0.847$  & $0.864$  & $0.0017$    & $\boldsymbol{0.0019}$  & $\boldsymbol{0.0424}$ \\
TwoRoom-W  & $0.1304$ & $0.0847$ & $3.76e{-4}$ & $\boldsymbol{0.0045}$ & $\boldsymbol{0.0783}$ \\
Reacher-W  & $0.0053$ & $0.0072$ & $8e{-5}$    & $\boldsymbol{0.0107}$  & $\boldsymbol{0.0909}$ \\
\bottomrule
\end{tabular}
\end{center}
\vspace{-0.6\baselineskip}
\end{table}

As summarized in Table~\ref{tab:disentangle}, the DWM pred head matches
the action sensitivity of the LeWM predictor in magnitude across all
three W-variants, while the world-head variance is significantly
reduced. The world/pred variance and range ratios remain well below
$10^{-2}$ and $10^{-1}$ across tasks, showing that the world head is
close to invariant to the action both on average and in the worst case,
while the pred head keeps the action-conditioned transition available
for planning.

\subsection{Multi-Step Prediction and OOD Dynamics Generalization}
\label{sec:results:multistep-ood}

Section~\ref{sec:results:evidence} verified that the decomposition
emerges at the representation level. We next ask whether it also
manifests in the learned latent dynamics, in two complementary ways:
by remaining stable when the source of the world effect is shifted
out of distribution, and by improving multi-step prediction along the
horizon that CEM actually queries.

\paragraph{OOD gravity generalization.}

\begin{wrapfigure}{r}{0.36\linewidth}
\vspace{-1.2\baselineskip}
\centering
\includegraphics[width=\linewidth]{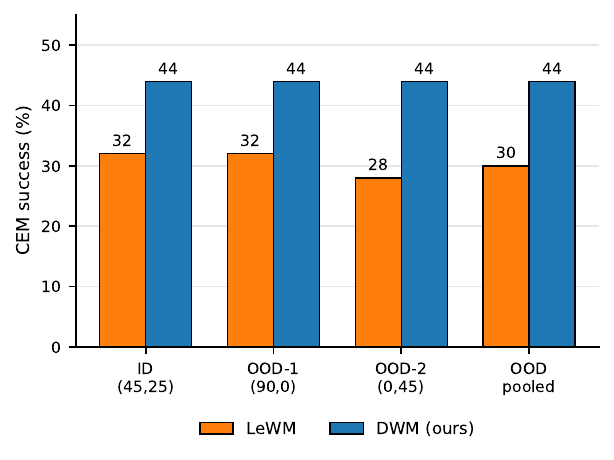}
\caption{OOD gravity CEM planning success on PushT-W.}
\label{fig:pusht-ood-gravity}
\vspace{-0.6\baselineskip}
\end{wrapfigure}
To test whether the learned decomposition further generalizes across
world dynamics, we train on PushT-W at gravity $(45,25)$ and evaluate at
held-out settings, comparing $\hat z^{\,a}$ across gravities and CEM
planning success under the protocol of Section~\ref{sec:results:main}.
Pooled over $(90,0)$ and $(0,45)$, the OOD-to-in-distribution
$\hat z^{\,a}$ cosine still reaches $0.9991$ (mean effect MSE $0.0015$),
indicating that DWM isolates a direction of action effects independent
of the gravity value. Control follows suit
(Fig.~\ref{fig:pusht-ood-gravity}): LeWM drops to $30.0\%$ pooled over
the two OOD gravities, while DWM retains $44.0\%$ ($+14$~pp), matching
the in-domain gap and thereby showing that the planning gains of
Section~\ref{sec:results:main} come from a gravity-agnostic
decomposition rather than overfitting to the training gravity.

\paragraph{Latent prediction and rollout quality.}
We evaluate the predictor on held-out validation windows of each
W-variant with one-step prediction and horizon-$20$ autoregressive
rollout. Error metrics follow Section~\ref{sec:setup:metrics}: latent
MSE for PushT-W and TwoRoom-W, and qpos error (rad) for Reacher-W.

\begin{table}[h]
\caption{Latent prediction quality on the three W-variant tasks.
\emph{Offline} is the one-step error on the held-out validation set;
\emph{Rollout@20} is the autoregressive error at horizon $20$. Error
metric: latent-space MSE for PushT-W and TwoRoom-W, qpos error (rad)
for Reacher-W.}
\label{tab:rollout}
\begin{center}
\small
\setlength{\tabcolsep}{4pt}
\begin{tabular}{llccc}
\toprule
\textbf{Task} & \textbf{Metric} & \textbf{LeWM} & \textbf{DWM} & \textbf{Relative change} \\
\midrule
PushT-W    & Offline \texttt{pred\_loss} $\downarrow$      & $0.372$ & $\boldsymbol{0.038}$ & $-89.8\%$ \\
PushT-W    & Rollout@20 \texttt{pred\_loss} $\downarrow$   & $1.034$ & $\boldsymbol{0.344}$ & $-66.7\%$ \\
\midrule
TwoRoom-W  & Offline \texttt{pred\_loss} $\downarrow$      & $0.317$ & $\boldsymbol{0.217}$ & $-31.6\%$ \\
TwoRoom-W  & Rollout@20 \texttt{pred\_loss} $\downarrow$   & $0.382$ & $\boldsymbol{0.261}$ & $-31.7\%$ \\
\midrule
Reacher-W  & Offline qpos error (rad) $\downarrow$         & $0.046$ & $\boldsymbol{0.033}$ & $-29.8\%$ \\
Reacher-W  & Rollout@20 qpos error (rad) $\downarrow$      & $0.299$ & $\boldsymbol{0.221}$ & $-26.1\%$ \\
\bottomrule
\end{tabular}
\end{center}
\vspace{-0.6\baselineskip}
\end{table}

Table~\ref{tab:rollout} shows that DWM reduces the one-step
prediction error on the three W-variants by $89.8\%$, $31.6\%$, and
$29.8\%$, and the horizon-$20$ error by $66.7\%$, $31.7\%$, and
$26.1\%$; on PushT-W its horizon-$20$ error ($0.344$) is even lower
than LeWM's one-step error ($0.372$). The planning gains of
Section~\ref{sec:results:main} thus rest on more accurate latent
dynamics along the horizons that CEM optimizes.

\subsection{Multiple Interacting World Effects: Ball-in-Cup}
\label{sec:results:hard}

We further evaluate DWM on Ball-in-Cup, where gravity,
string tension, and ball--cup contact jointly drive the ball's motion,
so a single trajectory couples several action-invariant physical
processes and probes DWM under multiple interacting world effects.
Under matched training and planning budgets, we compare LeWM and DWM
using CEM planning success and autoregressive prediction loss, with a
per-model success head scoring candidate trajectories
(Appendix~\ref{app:tasks}).

\begin{table}[h]
\caption{CEM planning success and autoregressive
prediction loss on Ball-in-Cup.}
\label{tab:bic-main}
\begin{center}
\small
\setlength{\tabcolsep}{4pt}
\begin{tabular}{lcc}
\toprule
\textbf{Model} & \textbf{CEM success} & \textbf{Rollout \texttt{pred\_loss}} $\downarrow$ \\
\midrule
LeWM  & $36.8\%$          & $0.671$ \\
DWM   & $\boldsymbol{42.8\%}$ & $\boldsymbol{0.089}$ \\
\bottomrule
\end{tabular}
\end{center}
\vspace{-0.6\baselineskip}
\end{table}

Table~\ref{tab:bic-main} shows that DWM raises CEM
planning success from $36.8\%$ to $42.8\%$ and cuts autoregressive
prediction loss from $0.671$ to $0.089$, confirming that the
world/action decomposition helps under multiple interacting world
effects.

\subsection{Generalization Across Distinct Latent World Models}
\label{sec:results:dinowm}

\begin{wrapfigure}{r}{0.36\linewidth}
\vspace{-1.2\baselineskip}
\centering
\includegraphics[width=\linewidth]{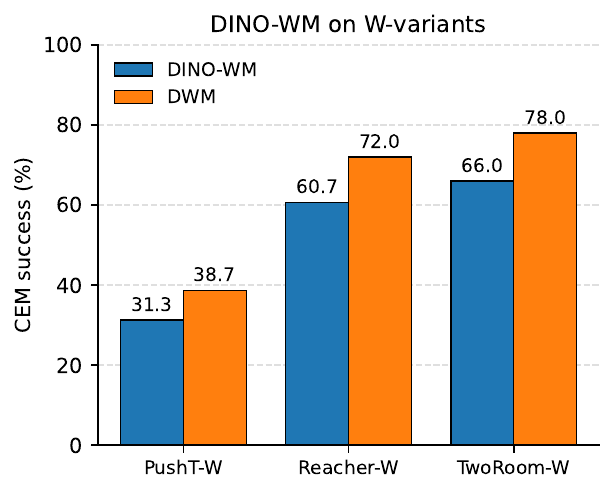}
\caption{Results on DINO-WM.}
\label{fig:dinowm-generalization}
\vspace{-0.6\baselineskip}
\end{wrapfigure}
To test whether DWM's benefits generalize beyond LeWM, we integrate the
same world/action decomposition into DINO-WM~\citep{zhou2024dino},
another representative action-conditioned latent world model that
operates on frozen DINOv2 patch features rather than a scratch-trained
encoder. Under matched training and CEM budgets, we compare planning
performance with and without DWM on the three W-variants
(Fig.~\ref{fig:dinowm-generalization}). Applied to DINO-WM, DWM raises
planning success on all three tasks by $10.2$ percentage points on
average. Together with the LeWM results, world/action
separation plugs cleanly into existing action-conditioned latent world
models and delivers consistent improvements across them.
\par

\section{Conclusion}
\label{sec:conclusion}

Joint-embedding latent world models are trained against a single
next-latent target, fusing action-driven and action-invariant changes
into one supervision signal and leaving the model unable to attribute
transitions to their causes---a supervision-level limitation that
undermines prediction and planning under realistic world effects. To
address this, we introduce DWM, a world/action-decomposed training
framework that supervises the two components separately through a
world head and an orthogonality constraint, delivering consistent
improvements across multiple more realistic benchmarks and multiple
representative action-conditioned latent world models.


\bibliography{references}
\bibliographystyle{iclr2026_conference}

\clearpage
\appendix

\section{Related Work}
\label{sec:related}

\paragraph{Latent world models for planning.}
Latent world models learn compact predictive representations for imagination, planning, and policy learning \citep{hafner2021dreamerv2,hafner2023dreamerv3,hansen2022tdmpc,hansen2024tdmpc2,maes2026lewm,micheli2023iris,robine2023twm}. Among them, joint-embedding world models have attracted growing attention for directly regressing future visual representations without reconstructing raw signals, yielding compact and semantically meaningful latents for downstream control. Representative works include LeWM~\citep{maes2026lewm} and DINO-WM~\citep{zhou2024dino}, whose transition prediction loss matches the combined next-state representation without explicitly allocating supervision between action-driven and action-invariant changes. Our method is complementary to prior advances in planners \citep{rubinstein1997cem,chua2018pets,hansen2024tdmpc2}, backbones \citep{bardes2024vjepa,caron2021dino}, and policy objectives \citep{hafner2020dreamer,hafner2023dreamerv3}. Instead, it modifies the supervision of a shared latent predictor by introducing two projections with complementary objectives, explicitly separating action-driven and action-invariant dynamics during training.

\paragraph{Decomposing latent transitions in world models.}
Most prior work disentangles \emph{what} a latent state represents rather than \emph{what causes its transition}: object-centric models factorize states into entities or spatial components~\citep{kipf2020cswm,ferraro2025focus,locatello2020slotattention}, while task-oriented methods split task-relevant from irrelevant factors~\citep{wang2022denoisedmdp,fu2021tia}. Closer to our goal, Iso-Dream~\citep{pan2022isodream} learns separate forward branches for controllable and noncontrollable dynamics, but this architectural split modifies the base world model and imposes a hard partition that is ill-suited when world and action effects act on the same object. NCL~\citep{wang2024ncl} enforces non-negativity for feature-level disentanglement, and PLSM~\citep{saanum2024softly} regularizes state-dependence of latent changes; neither, however, explicitly separates world from action effects, limiting their benefit when world effects are present. In contrast, DWM introduces an auxiliary world head with complementary training objectives to explicitly separate the two effects, improving action modeling and downstream planning while integrating easily with existing latent world models and preserving the original inference pipeline.

\section{Task and Dataset Details}
\label{app:tasks}

This appendix provides the full physical parameterization,
dataset scale, and out-of-distribution (OOD) splits for the tasks
introduced in Section~\ref{sec:tasks}. Table~\ref{tab:tasks}
summarizes the resulting benchmark suite; the paragraphs below
describe each pair in turn, together with Ball-in-Cup as
a more demanding simulated task with multiple
interacting world effects.

\begin{table}[h]
\caption{Control tasks used in this paper, grouped into three
flat/W-variant pairs (``-W'' for \emph{world effect}, and \emph{flat}
for the counterpart without the added persistent
world effect) plus Ball-in-Cup
as a more demanding simulated task with multiple
interacting world effects. Each W-variant introduces a persistent
action-invariant world effect; the flat counterpart shares the same
goal but removes that effect. Tasks in the same block share a goal
(block pushing, qpos matching, 2D target reaching, catching the ball
in the cup).}
\label{tab:tasks}
\begin{center}
\begin{tabular}{lll}
\toprule
\textbf{Task} & \textbf{Action-invariant world effect} & \textbf{Physical setting} \\
\midrule
PushT    & None                          & Horizontal table \\
PushT-W     & Gravity-driven sliding        & Gravity $(45,25)$, damping $0.92$ \\
\midrule
Reacher  & None                          & Standard DMC Reacher \\
Reacher-W   & Vertical-plane gravity        & $g{=}{-}9.81\sin 60^\circ$ along $z$ \\
\midrule
TwoRoom  & None                          & $\Delta p{=}a\cdot s$ \\
TwoRoom-W   & Constant drift bias           & $\Delta p{=}a\cdot s + b$, $b{=}(4,2)$ \\
\midrule
Ball-in-Cup     & Multiple interacting effects & Standard DMC Ball-in-Cup \\
\bottomrule
\end{tabular}
\end{center}
\end{table}

\paragraph{PushT / PushT-W.} 
Standard PushT \citep{chi2023diffusion} is a horizontal-table pushing task: the block does not drift
under zero action, so action-invariant world effects are weak. Its
W-variant counterpart keeps the same pushing objective but includes gravity
$(45,25)$ in the \texttt{pymunk} space (with damping $0.92$), so the
block continues to slide even when the agent is idle. We use standard
PushT as the flat comparability control, and train PushT-W on
\texttt{tilted\_pusht\_train\_4x} with 3200 episodes $\times$ 200 steps
under a random policy; for OOD stress tests,
we additionally evaluate at gravity $(90,0)$ and $(0,45)$
(Section~\ref{sec:results:multistep-ood}). CEM planning uses a goal
offset of $10$ frames, an evaluation budget of $50$ environment
steps, and a CEM configuration of
$300\!\times\!30\!\times\!\mathrm{top}30$ ($300$ samples, $30$
refinement iterations, top-$30$ elite set); an attempt is counted
as a success if the block matches the goal configuration at the
final step of the budget.

\paragraph{Reacher / Reacher-W.}
Standard DMC Reacher \citep{tassa2018dmc,tunyasuvunakool2020dmcontrol} operates on a horizontal plane, where action-invariant world effects
are negligible.
Reacher-W is a customized DMC Reacher in
which the shoulder and wrist hinge axes are rotated from
$[0,0,1]$ to $[0,1,0]$ so that the arm moves in the vertical $x$--$z$
plane while keeping the same qpos-matching objective. Under the
standard gravity vector
$g\!=\!(0,0,-9.81\sin 60^\circ)\!\approx\!(0,0,-8.50)$~m/s$^2$,
gravity therefore exerts a non-zero torque on the shoulder joint and
drives passive motion of the arm even under a null action, whereas in
the original horizontal geometry the same gravity vector is parallel
to the shoulder axis and produces no such torque. Here the world
effect refers to this gravity-driven passive motion under a null
action, not an additional external force channel.
Both variants use $10{,}000$ episodes $\times\,200$ steps, with a held-out
validation set of $500$ episodes $\times\,200$ steps generated under
the same physics and policy mix. CEM planning uses a goal offset of
$25$ frames, an evaluation budget of $200$ environment steps, and a
CEM configuration of $300\!\times\!10\!\times\!\mathrm{top}30$; an
attempt succeeds if the joint configuration reaches within
$0.15$~rad of the goal qpos at any step within the budget.

\paragraph{TwoRoom / TwoRoom-W.}
Standard TwoRoom is a 2D navigation environment with transition
$p_{t+1}=p_t+a_t\cdot s$, so the agent does not drift under zero action
and action-invariant world effects are absent. TwoRoom-W keeps the
same reach-target objective but adds a persistent environmental drift,
$p_{t+1}=p_t+a_t\cdot s + b$ with $b=(4,2)$ pixels/step. We train the
W-variant on 2000 episodes $\times$ 200 steps with a random policy
under the same drift, yielding $324{,}498$ training windows and a
matched validation set of $60\!\times\!100$ transitions.
CEM planning on TwoRoom-W adopts a longer-horizon CEM-MPC protocol
to accommodate the persistent drift: goal offset $90$ frames,
evaluation budget of $25$ MPC decisions, each planning over a
rollout horizon of $5$ steps and executing an action block of $5$
environment steps before re-planning, with the same
$300\!\times\!10\!\times\!\mathrm{top}30$ CEM budget as Reacher-W;
success is declared when the agent reaches the target region within
the total budget.

\paragraph{Ball-in-Cup.}
The DMC Ball-in-Cup task
\citep{tassa2018dmc,tunyasuvunakool2020dmcontrol} requires a movable cup
to toss and catch a ball attached to it by a string. The training set
contains $800$ episodes of $200$ steps each. Both world models are
trained for $20$ epochs with the configuration in
Appendix~\ref{app:impl}.

After world-model training, we fit a separate success
head to each model's autoregressively predicted latents, supervised by
success labels from the corresponding environment trajectories. Each
model uses its own success head to rank candidate action sequences
during CEM planning. For DWM, this head receives the complete
predictions from the pred head; the auxiliary world head is used only
during world-model training. Both models use the same CEM budget.
Planning success measures whether the ball is caught within the
evaluation budget, and autoregressive prediction loss is evaluated over
the planning horizon.
Planning success is averaged over $100$ start states
per seed and reported as the mean over seeds $\{42,43\}$. CEM uses a
rollout horizon of $5$ frames with a receding horizon of $5$ and an
action block of $5$ environment steps, an evaluation budget of $50$
environment steps, and a per-decision budget of $150\!\times\!10\!
\times\!\mathrm{top}15$ samples with residual variance $0.08$. Each
success head is a two-layer MLP trained for $30$ epochs on $8{,}000$
rollout sequences generated by the corresponding world model at the
same horizon, with success labels taken from the underlying Ball-in-Cup
trajectories; the two models are compared under matched training and
planning protocols.

\section{Implementation Details}
\label{app:impl}

Both base models share the DWM additions of Section~\ref{sec:method}: a
world head $h_w$ attached in parallel to the pred head $h_p$, an
action-shuffle world-contrastive loss $\mathcal{L}_{\mathrm{wc}}$, and an
orthogonality penalty $\mathcal{L}_{\mathrm{orth}}$ between the world and
action-residual directions, with contrastive temperature $\tau=0.07$ and
loss weights $\lambda_{\mathrm{wc}},\lambda_{\mathrm{orth}}$ from
Eq.~\ref{eq:method_loss}. At inference the world head is discarded and
each base model's original prediction and planning pipeline is used
unchanged. Below we give the per-model configuration.

\paragraph{LeWM.}
The encoder $\varphi$ is a ViT-tiny (hidden dimension $192$, $12$ layers,
patch size $14$, input resolution $224^2$) trained from scratch, and the
action-conditioned predictor $g$ is a $6$-layer Transformer ($16$
attention heads, head dimension $64$, MLP dimension $2048$) with AdaLN
action conditioning; the context length is $h\!=\!3$ and observations are
subsampled at a stride of $5$ frames. The pred head $h_p$ and world head
$h_w$ are each a two-layer MLP
$192\!\rightarrow\!2048\!\rightarrow\!192$ with BatchNorm. The encoder,
predictor, and both heads are initialized randomly and optimized jointly
with AdamW ($\mathrm{lr}\!=\!5\!\times\!10^{-5}$ across all modules,
weight decay $10^{-3}$) under a linear-warmup cosine-annealing schedule
at the epoch level, in \texttt{bfloat16} on NVIDIA H20 GPUs with batch
size $128$. The number of epochs is fixed per task family before
evaluation: $5$ for PushT / PushT-W, $20$ for Reacher / Reacher-W, $100$
for TwoRoom / TwoRoom-W, and $20$ for Ball-in-Cup.

\paragraph{DINO-WM.}
The encoder $\varphi$ is a frozen pretrained DINOv2 ViT-S/14
(\texttt{dinov2\_vits14}, patch size $14$), producing $384$-dimensional
patch features, and the predictor $g$ is DINO-WM's $6$-layer causal ViT
predictor over the $P$ patch tokens per frame, with the same context
length and frame stride as above. The world head $h_w$ is a token-wise
two-layer MLP $384\!\rightarrow\!2048\!\rightarrow\!384$ with BatchNorm,
applied independently to each patch token, while the pred head $h_p$ is
the identity on $r_t$, so the predicted next latent $\hat z\!:=\!r_t$
matches the baseline and only the world head is added. The main loss
$\mathcal{L}_{\mathrm{WM}}$ is DINO-WM's original prediction and
reconstruction objective, and both auxiliary losses use the same
formulation as for LeWM, applied to the visual portion of the token
embedding and averaged over (time, patch) locations. The predictor,
action/proprio encoders, decoder, and world head are optimized jointly
under the base model's schedule with batch size $256$ for PushT-W and
Reacher-W and $128$ for TwoRoom-W on $8$ NVIDIA H20 GPUs, with the epoch
budget fixed per task family before evaluation. At inference the world
head is discarded and the raw predictor output is fed to the CEM planner
as in the baseline.
\par

\section{Ablation Studies}
\label{app:ablation}

We conduct ablation studies on PushT-W to
evaluate the contributions of $\mathcal{L}_{\mathrm{wc}}$ and
$\mathcal{L}_{\mathrm{orth}}$ to DWM. We vary each loss weight while
holding the other fixed and measure CEM planning success.

\begin{table}[h]
\caption{Loss-weight ablation on PushT-W
(CEM planning success, $n{=}50$). The weight pair $(0,0)$ corresponds
to the LeWM baseline; other configurations vary one weight while
holding the other fixed.}
\label{tab:ablation-pusht-grid}
\begin{center}
\small
\setlength{\tabcolsep}{4pt}
\begin{tabular}{lcccccc}
\toprule
$\lambda_{\mathrm{wc}}$
& $0.0$
& $0.1$
& $0.2$
& $0.3$
& $0.3$
& $0.3$ \\
$\lambda_{\mathrm{orth}}$
& $0.0$
& $0.5$
& $0.5$
& $0.0$
& $0.1$
& $0.5$ \\
\midrule
\textbf{CEM success}
& $32.0\%$
& $36.0\%$
& $38.0\%$
& $40.0\%$
& $42.0\%$
& $\boldsymbol{44.0\%}$ \\
\bottomrule
\end{tabular}
\end{center}
\vspace{-0.6\baselineskip}
\end{table}

Along the world-contrastive axis, reducing
$\lambda_{\mathrm{wc}}$ from $0.3$ to $0.2$ to $0.1$ while keeping
$\lambda_{\mathrm{orth}}=0.5$ yields CEM success of
$44.0\%\!\to\!38.0\%\!\to\!36.0\%$: as the contrastive signal weakens,
the world head deviates from action-invariance and forfeits most of
the gain over the LeWM baseline. Along the orthogonality axis,
reducing $\lambda_{\mathrm{orth}}$ from $0.5$ to $0.1$ to $0.0$ while
keeping $\lambda_{\mathrm{wc}}=0.3$ gives
$44.0\%\!\to\!42.0\%\!\to\!40.0\%$, showing that an action-invariant
world head alone does not couple its component into $\hat z$ tightly
enough to close the remaining gap. DWM attains its best CEM success
($44.0\%$) only when $\mathcal{L}_{\mathrm{wc}}$ and
$\mathcal{L}_{\mathrm{orth}}$ are jointly active, confirming that both
losses are integral to the framework.

\section{Planning Success on Flat Counterparts}
\label{app:flat-suite}

This appendix reports the behavior of DWM
on the flat counterparts of the three W-variant tasks
(Section~\ref{sec:results:main}), where the environment does not
inject a persistent action-invariant world effect. All models share
the training recipe of Section~\ref{sec:setup:model} and the CEM
protocol of Section~\ref{sec:prelim:cem}; means and standard
deviations are computed over $3$ CEM planner seeds with $n{=}50$
start--goal pairs per seed.

\begin{figure}[h]
\centering
\includegraphics[width=0.7\linewidth]{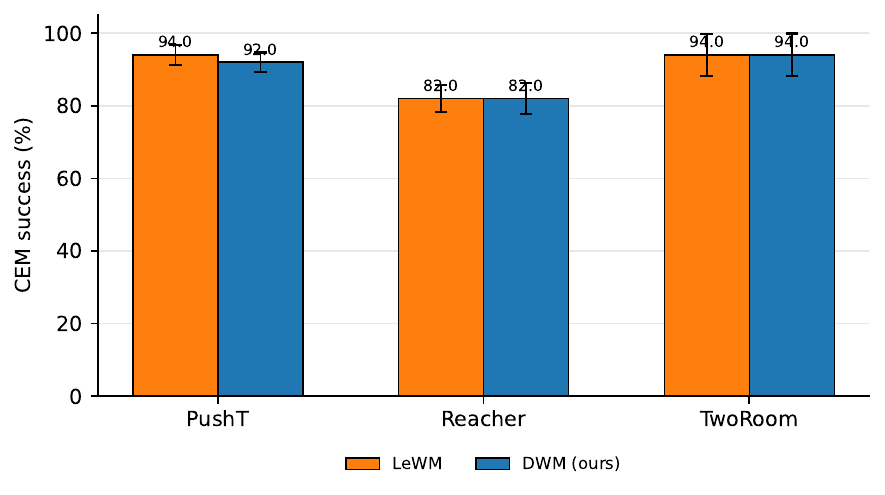}
\caption{CEM planning success rate on the
flat counterparts of PushT, Reacher, and TwoRoom (no persistent
action-invariant world effect). DWM stays within noise of the LeWM
baseline on all three tasks, indicating that the auxiliary
supervision does not harm planning when the world component it
targets is absent. Same training recipe and CEM configuration as
Section~\ref{sec:results:main}, averaged over $3$ CEM planner seeds
($n{=}50$ start--goal pairs per seed); error bars denote
across-seed standard deviation.}
\label{fig:flat-suite}
\end{figure}

\end{document}